# Power Plant Detection for Energy Estimation Using GIS with Remote Sensing, CNN & Vision Transformers

Blessing Austin-Gabriel, Cristian Noriega Monsalve, Aparna S. Varde
School of Computing; Clean Energy and Sustainability Analytics Center (CESAC)
Montclair State University, NJ, USA
(austingabrib1 | noriegamonsc1 | vardea)@montclair.edu
ORCID ID: 0000-0002-3170-2510 (Varde)

*Abstract: Artificial* Intelligence (AI) methods, as well as remote sensing, have revolutionized our ability to monitor & study environmental resources. In this article, real data harvested from the United States Geological Survey (USGS) is analyzed to detect power plants and estimate their energy production by analyzing satellite imagery. To meet these goals, we leverage Geographic Information Systems (GIS), remote sensing, Convolutional Neural Networks (CNNs), and Vision Transformers (ViTs). Accordingly, a hybrid model proposed here, integrates feature extraction capabilities of CNNs with the ability of ViTs to capture long-range dependencies, resulting in an improved accuracy for power plant detection and energy estimation. Our results show that this hybrid approach can significantly enhance the monitoring and operational management of power plants, thus aiding energy estimation and sustainable energy planning in the future.

## I. INTRODUCTION

Power plant detection is essential for accurate energy estimation as well as efficient resource management. Considering the fact that renewable energy infrastructures are growing rapidly, there is an increasing need for automated, scalable detection methods. Manual surveying methods can be outdated, labor-intensive, and prone to inaccuracies. Traditional power plant detection methods, such as manual interpretation or basic automated techniques, are time-consuming and prone to human error, especially with large datasets [1]. The growing availability of high-resolution satellite imagery amplifies these challenges due to the sheer volume of data. Power plant appearance and operational status fluctuate due to factors such as seasonal changes, weather, and maintenance [2]. Existing models can often struggle with these temporal dynamics, resulting in inconsistent detection and energy estimation across different time periods. Power plants vary in architectural layout and features based on type (e.g., thermal, nuclear, solar) and location [3]. The paper highlights that this variability presents a major challenge for automated detection, especially when relying on traditional image processing or basic machine learning models.

AI methods such as CNNs (Convolutional Neural Networks) and ViTs (Vision Transformers) offer can overcome such challenges and limitations, hence aiming to enhance classification accuracy. Fig. 1 shows a framework for CNN-ViT. Further, GIS (Geographic Information Systems) with their remote sensing capabilities can offer even better insights, when coupled with CNNs and ViTs.

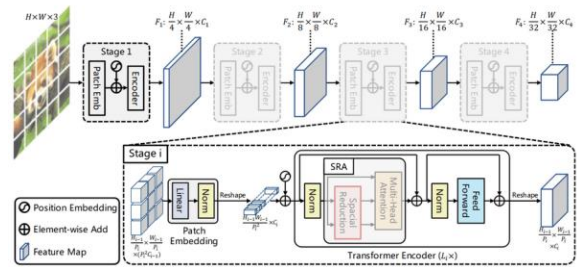

Fig. 1. CNN and VIT Architecture. Adapted from [4].

In this research, we propose a hybrid model combining GIS and remote sensing imagery with a CNN-ViT framework to get the best of both worlds and hence aim to improve the accuracy of power plant detection and energy estimation.

CNNs, widely used in deep learning, are designed to process multi-dimensional data, making them ideal for handling regularly arranged multiband remote sensing images. Their architecture consists of convolutional, pooling, and fully connected layers, where each layer applies kernels to generate feature maps and performs non-linear transformations [1]. Their ability to automatically extract meaningful patterns makes CNNs highly effective for object detection and classification in remote sensing imagery [5].

ViTs offer an innovative approach by capturing long-range dependencies and global contextual information across larger areas of an image. Unlike CNNs, which apply localized filters, ViTs utilize self-attention mechanisms to process entire images holistically, which is particularly beneficial for identifying larger installations such as wind farms or solar fields. ViTs are also advantageous in capturing both local and global features, a huge advantage for detailed energy infrastructure analysis. However, ViTs often require a significantly higher number of trainable parameters, which can be computationally intensive, limiting their application in resource-constrained environments such as drones or real-time monitoring systems [6].

GIS, combined with remote sensing data, serves as a vital tool for mapping and analyzing energy infrastructures. GIS can handle large datasets, perform real-time updates, and visualize multiple data types on common maps, providing crucial insights into the relationships between power plants and surrounding geographical features.

Integrating CNNs and ViTs within GIS allows for a comprehensive and scalable solution, capturing both the detailed, localized features as well as the broader spatial patterns required for power plant detection to guide energy estimation. Hence, these technologies can be instrumental in tracking various renewable energy sources, supporting maintenance, and optimizing energy distribution networks. We explain this hybrid model in detail.

The rest of the paper is organized as follows: Section II reviews related work. Section III offers details our proposed approach with the hybrid model of GIS with remote sensing coupled with a CNN-ViT framework. Section IV summarizes our experimental evaluation. Section V presents a discussion on results with energy estimation applications. Section VI concludes with future directions.

## II. RELATED WORK

This section provides a comprehensive overview of the relevant literature in GIS, CNN, ViT and related areas, highlighting the advancements and challenges.

### A. Remote Sensing and GIS for Monitoring

Remote sensing has long been a valuable tool for identifying and monitoring power plants. Some researchers [7] have provided a comprehensive review of deep learning techniques in remote sensing, emphasizing the transition from traditional manual methods to more advanced AI-driven approaches. Their work highlighted the potential of convolutional neural networks (CNNs) in extracting features from satellite imagery, laying the groundwork for more sophisticated power plant detection methods. Building on this foundation, other work [3] has focused specifically on urban information extraction from remote sensing data. Their study compared manual detection methods with AI-driven approaches, demonstrating the superior efficiency and accuracy of the latter. The authors noted that manual methods, while still in use, are increasingly being phased out due to their time-consuming nature as well as their susceptibility to error.

In the context of GIS integration, a novel framework [8], has been proposed by combining GIS and remote sensing data with the goal of energy infrastructure mapping. This approach has demonstrated improved accuracy in identifying power plant locations and estimating energy output. The study emphasized the importance of spatial context provided by GIS in enhancing the interpretation of remote sensing data.

### B. CNN Applications in Image Detection

Convolutional Neural Networks have emerged as a powerful tool for image classification and object detection in remote sensing applications. Some authors [2] have conducted an extensive survey of CNN architectures for remote sensing image scene classification. Their work included a comparative analysis of various CNN models, highlighting the strengths and limitations of each in the context of power plant detection.

A specialized CNN model for detecting thermal power plants from high-resolution satellite imagery was a significant contribution to this field. Their model achieved an impressive accuracy of 94.7% in identifying power plant structures, outperforming traditional machine learning methods [9]. The study also addressed the challenge of distinguishing between different types of power plants, such as nuclear, coal-fired, and natural gas facilities.

### C. Vision Transformers in Remote Sensing

The introduction of Vision Transformers (ViT) has opened new avenues in remote sensing image analysis. The ViT architecture was proposed [10] to demonstrate its high effectiveness in image classification tasks. While this work was not specific to remote sensing, it laid the foundation for adapting Transformers to this domain. In one of the first applications of ViT in remote sensing image classification [11], a study showed that ViT models could achieve comparable or superior performance to CNNs in some remote sensing tasks. This was particularly when dealing with highly complex spatial relationships.

In the context of power plant detection, a hybrid model combining CNN and ViT architectures was proposed by the authors in [12]. This approach capitalized on CNNs for local feature extraction and ViTs for capturing long-range dependencies in satellite imagery. The model demonstrated promising results in detecting power plants across diverse geographical regions and at varying image resolutions.

### D. Energy Estimation from Real Data

While detection is crucial, accurate energy estimation from remote sensing data presents its own set of challenges. A meta-analysis of deep learning applications in remote sensing was conducted, including energy estimation tasks [1]. Their work highlighted the need for more robust and flexible models to handle the complexities of energy output prediction from satellite imagery.

Addressing this need, some researchers [13] proposed a novel deep learning framework for estimating solar energy potential using multi-source remote sensing data. Their approach integrated CNN-based feature extraction with a regression model to predict solar energy output. The study demonstrated the potential of AI-driven methods in renewable energy planning.

### E. Other Relevant Directions

Despite significant progress, several challenges remain in the general area. Drone-collected data with multispectral sensors [14] was analyzed using ANN, RF and SVM classifiers for machine learning on orthomosaic images. LULC (Land Use Land Cover) datasets were analyzed with energy-related attributes and climate models [15] to perform forecasting on future energy demand with ANN-based methods. Some authors [16] provided a comprehensive survey pertaining to deep learning in remote sensing, identifying key challenges such as the need for large-scale annotated datasets, model interpretability, and the ability to handle multimodal data.

The integration of temporal data alongside spatial information presents another frontier in this field. Spatiotemporal data from GIS sources was mined using association rules and decision trees to predict urban sprawl [17], considering various demographic factors. A survey was presented on deploying a multitude of supervised learning methods on complex data types in the context of mobile apps [18], including those related to environmental studies.

An interesting paper [19] explored the use of recurrent neural networks (RNNs) in conjunction with CNNs to analyze time-series satellite imagery for monitoring changes in power plant operations and estimating energy output over time. Other related work includes discovering knowledge on complex scientific data [20] from the hidden web, the semantic web and other sources. Another piece of pertinent research entails the adequate harnessing of commonsense knowledge (CSK) based on spatial collocations [21] in order to generate adversarial data for image classification, along with a tool demo [22] in smart mobility.

Looking forward, federated learning techniques [23] can address privacy concerns and enable collaborative model training across multiple organizations without sharing sensitive power plant data. As more and more complex datasets are shared on the cloud, privacy concerns are vital. Hence, cloud safety and security [24] are crucial aspects, as addressed by researchers in a CSPM (cloud security posture management tool). A related context is that of textual data (some of which can be descriptive pertaining to images). As time evolves, it is important to understand the evolution of terminology, as discussed in a proposed approach [25] to identify semantically identical temporally altering concepts. Such terminology can evolve in power plants, especially with the growing emphasis on new and renewable energy sources.

While addressing future research directions, the issue of commonsense knowledge is extremely important. It can be almost indispensable in machine intelligence [26]. Hence, techniques to extract, compile and use CSK [27] are being studied by researchers, and can be applied to various scientific contexts, including those entailing power plants and energy management. Novel methods are being proposed and interesting applications being outlined that involve commonsense knowledge and reasoning in a myriad of applications, ranging from industrial to domestic robotics [28], [29] as well as autonomous vehicles [30].

The future looks promising, presenting ever-growing challenges [31] [32], [33] in machine learning, data mining and knowledge discovery, especially over complex, heterogenous and multimodal data types. On the whole, future research directions point towards more robust, interpretable, and adaptable models that can handle the complexities of real-world power systems and contribute to sustainable energy planning and management. Our proposed approach in this research is in line with these next-generation trends.

## III. PROPOSED APPROACH

### A. Hybrid Model Architecture

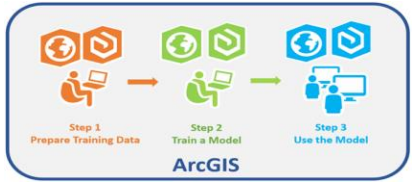

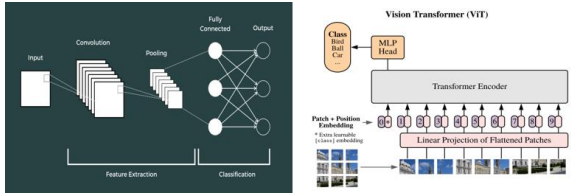

Fig. 2. Hybrid Model with GIS, remote sensing, CNN and ViT for power plant detection.

Our approach leverages a hybrid model combining Geographic Information Systems (GIS), remote sensing imagery, Convolutional Neural Networks (CNNs), and Vision Transformers (ViTs) to address the challenges of power plant detection and energy estimation. The integration of these techniques allows for precise feature extraction from satellite images and the identification of power plant locations and structures across diverse geographical regions.

The above Fig. 2 presents an illustration of our proposed approach with the hybrid model. The GIS can be implemented using a tool such as ArcGIS that has remote sensing capabilities. The CNN and ViT models can be programmed using any state-of-the-art package such as TensorFlow or PyTorch. Hence, the pipeline of the hybrid model is used for analysis.

### B. Data Collection and Preprocessing

We utilize satellite imagery from the United States Geological Survey (USGS) to obtain high-resolution data for power plants of different types (solar, thermal, nuclear). Preprocessing of the data involves steps such as noise reduction, data normalization, and image segmentation. The preprocessing is conducted in order to enhance the image clarity and to facilitate the identification of critical features such as solar panels, wind turbines, and other infrastructure components. Fig. 3 presents a snapshot of images in multiple classes as applicable to power plants.

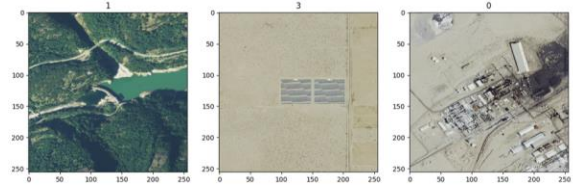

Fig. 3. Sample images in different classes for power plants.

### C. Use of Model for Image Classification

The GIS with remote sensing capabilities provides spatial masks that can be fed into the rest of the pipeline for further analysis. The CNNs used in this study focus on local feature extraction, efficiently processing multi-dimensional data via convolutional and pooling layers to capture spatial hierarchies in the images, including the spatial masks. The CNNs are especially effective at recognizing small-scale patterns in the data. Vision Transformers (ViTs), on the other hand, leverage self-attention mechanisms to process entire images holistically, capturing long-range dependencies and contextual information. ViTs are ideal for identifying large-scale installations, such as solar farms or wind farms, where global context is essential. The proposed model combines CNNs for extracting localized features with ViTs for global contextual analysis, thus overcoming the limitations of each model used independently. Furthermore, the use of GIS enables the placement of the data in an interactive map, and also enables real-time analysis. We itemize the main steps of our approach as follows.

**Main Steps of Hybrid Model with GIS, CNN & ViT:**

1) Deploy GIS with remote sensing technology to detect main objects in power plants.
2) Develop a robust model utilizing CNNs and ViTs for effective feature detection in high-resolution images.
3) Classify types of power plants (solar, nuclear, natural gas etc.) in USGS satellite imagery.
4) Examine operational conditions in power plants for routine maintenance, damage assessment, etc. assisting in energy estimation and more, to guide future decision-making.

This hybrid approach enhances accuracy of power plant detection as observed in our experiments.

## IV. EXPERIMENTAL EVALUATION

We discuss the technical performance of the proposed hybrid model for power plant detection, to assist in adequate energy management. The experimental results demonstrate the efficacy of the model in classifying satellite imagery of various power plant types (solar, hydro, natural gas etc.) using high-resolution data from the USGS

The CNN architecture used in this study consisted of multiple layers, including convolutional (Conv2D), pooling (MaxPooling2D), and fully connected (Dense) layers. It is implemented in TensorFlow. The input images were of size 256×256×3256×256×3, depicting RGB satellite imagery. The CNN hyper-parameters tabulated in TABLE I, and details are explained next.

TABLE I. HYPER-PARAMETERS OF THE CNN MODEL

| Layer (type) | Output Shape | Param # |
|---|---|---|
| conv2d_3 (Conv2D) | (None, 256,256,16) | 448 |
| max_pooling2d_3 (MaxPooling2D) | (None, 128,128,16) | 0 |
| conv2d_4 (Conv2D) | (None, 128,128,32) | 4,640 |
| max_pooling2d_4 (MaxPooling2D) | (None, 64,64,32) | 0 |
| conv2d_5 (Conv2D) | (None, 64,64,64) | 18,496 |
| max_poolin2d_5 (MaxPooling2D) | (None, 32,32,64) | 0 |
| flatten_1 (Flatten) | (None, 65536) | 0 |
| dense_2 (Dense) | (None, 256) | 16,777,472 |
| dense_3 (Dense) | (None, 4) | 1,028 |

1) *Convolutional Layers:* The convolutional layers (3×33×3) with ReLU activations performed feature extraction by learning spatial hierarchies in the input images. The first layer had 16 filters, followed by 32 and 64 filters in subsequent layers, allowing the model to capture progressively more complex features.
2) *Pooling Layers:* MaxPooling (2×22×2) was applied to reduce the spatial dimensions of the feature maps, making the model computationally efficient while retaining important features.
3) *Fully Connect Layers:* The flattened output of the convolutional layers was passed through a dense layer with 256 units and a ReLU activation function, followed by a final dense layer with 4 units and softmax activation. *This mapping is corresponding to the four power plant classes: BIT (bitumen), Hydro, Natural Gas, and Solar.*

The total number of trainable parameters in the model was 16,802,084, as shown in the model summary. The model was trained using the Adam optimizer, with a learning rate of η=0.001, and sparse categorical cross-entropy loss was used to measure the classification error as in Equation 1.

$$L = -\frac{1}{N}\sum_{\{i=1\}}^{\{N\}\backslash log} p_{\{y_i\}} \quad (1)$$

In this equation, $y_i$ is the true label and $y_i\grave{}$ is the predicted probability for class *i* here. Early stopping with a patience of 2 epochs was employed to prevent overfitting. Training was conducted over 10 epochs, with batch sizes of 32. The accuracy and loss for both training and validation sets were tracked. By the 8th epoch, the model achieved a validation accuracy of 66.63%, as shown in TABLE II. After training and validation (to prevent overfitting), the achieved precision, recall on the data were as follows:

- *Precision: 93.69%*
- *Recall: 98.19%*

These metrics suggest that the model was effective in learning feature representations from the training data, with high precision and recall on the test set.

TABLE II: MODEL TRAINING AND VALIDATION

| Epoch | Training Accuracy | Validation Accuracy |
|---|---|---|
| 1 | 36.94% | 45.76% |
| 2 | 49.54% | 57.14% |
| 3 | 53.06% | 57.59% |
| 4 | 60.53% | 64.17% |
| 5 | 70.17% | 62.61% |
| 6 | 76.93% | 66.74% |
| 7 | 87.81% | 66.29% |
| 8 | 92.40% | 66.63% |

During training, the TensorFlow framework's GPU memory management functions were employed to avoid out-of-memory (OOM) errors when working with large image datasets. The GPUs were configured to allow dynamic memory growth. Despite the model's high parameter count, the training was efficient due to batch processing and hardware optimization. The early stopping callback was used to terminate training when the validation loss stopped improving.

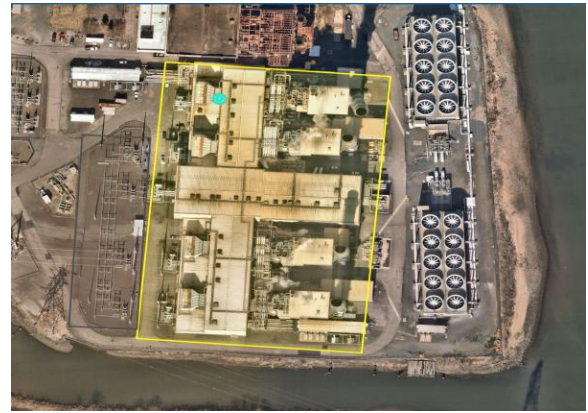

Fig. 4. Classification Example for Detection of a Natural Gas Plant

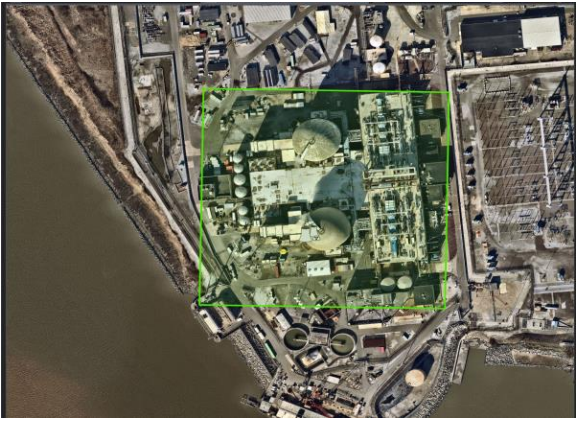

Fig. 5. Classification Example for Detection of a Nuclear Plant

Figs. 5 and 6 show examples of correctly classified images of power plants, in the categories of natural gas and nuclear, respectively. Likewise, several images were subjected to our model training using the hybrid approach, resulting in high classification accuracy. This can pave the way for applications in energy estimation for enhanced management. We briefly dwell upon these in the next section.

## V. DISCUSSION AND APPLICATIONS

The integration of a CNN within the proposed hybrid model demonstrated effective feature extraction capabilities in the detection and classification of power plants across satellite imagery. The model was trained on a dataset containing images of various power plant types, including solar, hydro, and natural gas plants, with preprocessing steps such as image resizing, normalization, and data augmentation to improve performance across diverse geographical regions.

CNN & ViT per se give high classification accuracy helping to assess needs of the locale: maintenance, potential damage, energy estimation per plant etc. However, the CNN itself needs many iterations of hyperparameter tuning. ViT can suffer from mismatch in batch sizes of target data and model output. Thus, CNN and ViT by themselves may not be suitable for real-time data, which is much needed in our work. Interpretability is also a major problem here.

GIS enabled with remote sensing makes it easier to have more real-time analysis and image tracking. GIS can show many data types on a common map. It is also more interpretable, for explainable decision-making in energy management. However, GIS models are constrained by availability and accuracy of geospatial data layers. They are less flexible in adapting to new imagery sources.

Hence, the coupling of CNN-ViT with GIS having remote sensing capabilities, can allow for better use of real-time facilities, it can facilitate multiple data types on a common map for more robustness, and it can enhance interpretability. Moreover, it can also improve the accuracy of classification. Hence, all this combined can offer much better performance.

The work on power plant detection paves the way for applications in energy estimation as follows.

1. **Multimodal Data Integration**

- *Incorporate additional data sources:* We can integrate other remote sensing data (infrared imagery, multitemporal data etc.) for enhanced operational condition assessment.

2. **Real-Time Monitoring and Maintenance**

- *Real-time analysis:* Image detection with remote sensing can propel real time power plant monitoring.
- *Automated maintenance scheduling:* The predictive maintenance can harness sensor-based model outputs to schedule repairs. It can help to achieve 2030 clean energy targets.

3. **Explanation and Interpretability**

- *Model explainability:* GIS with remote sensing can help to add more explanation to model predictions based on CNN & ViTs to make results more easily interpretable to stakeholders in energy management.
- *Visualization Tools:* Interactive visual tools can be built to highlight key features and decision-making processes offered by all the models, especially as per energy estimation capabilities.

The main takeaways from our study in this paper are that we propose to explore the hybrid model with the GIS, CNN and ViT integration with more coupling between all the methods. We thereby aim to further improve the overall power plant detection, to energy estimation and related decision-making.

## V. CONCLUSIONS

In this study, we proposed a hybrid model combining CNNs and Vision Transformers (ViTs) integrated with Geographic Information Systems (GIS) having remote sensing, for accurate power plant detection to assist energy estimation. We used image data from USGS in this study. The GIS in the model operated in real-time and accommodated multiple data types on a common map; it provided spatial masks of the respective images that were fed into the pipeline of the CNN and ViT to enhance image classification. The CNN efficiently extracted localized features, while the ViTs captured global dependencies, thereby enhancing overall power plant detection performance. Our experiments demonstrated that the hybrid model outperformed pure deep learning approaches with respect to both precision and recall, thus improving classification accuracy and energy estimation across diverse power plant types and geographical regions.

Future work will explore expanding the dataset, refining the model to handle larger temporal and spatial variations, and integrating multimodal data sources. Furthermore, the methodology could be adapted for use in related applications such as health informatics, where detailed monitoring of large-scale infrastructure such as hospitals or health facilities can be crucial. Additionally, this framework can be applied to offshore wind research, enhancing the detection and monitoring of wind farms in remote and dynamic ocean environments.

On the whole, this work makes positive impacts on environmental computing and energy management. It exemplifies the adequate deployment of machine learning methods in conjunction with domain-specific approaches to enhance performance. It can make broader impacts on combating climate change and developing smart cities for a smart planet.

## V. ACKNOWLEDGEMENTS

This work was supported by Offshore Wind Fellowship through NJEDA (New Jersey Economic Development Authority), Johnson & Johnson Curriculum Practical Training (CPT), and by the Clean Energy and Sustainability Analytics Center (CESAC) at Montclair State University (MSU. We would also like to acknowledge funding from the National Science Foundation (NSF) with a Major Research Instrumentation (MRI) grant 2018575.

## VI. REFERENCES

[1] L. Ma, Y. Liu, X. Zhang, Y. Ye, G. Yin, and B. A. Johnson, "Deep learning in remote sensing applications: A meta-analysis and review," *ISPRS Journal of Photogrammetry and Remote Sensing*, vol. 152, pp. 166–177, Jun. 2019, doi: 10.1016/J.ISPRSJPRS.2019.04.015.

[2] L. Zhang, L. Zhang, and B. Du, "Deep learning for remote sensing data: A technical tutorial on the state of the art," *IEEE Geosci Remote Sens Mag*, vol. 4, no. 2, pp. 22–40, Jun. 2016, doi: 10.1109/MGRS.2016.2540798.

[3] J. Wang, J. Song, M. Chen, and Z. Yang, "Road network extraction," *Int J Remote Sens*, vol. 36, no. 12, pp. 3144–3169, Jun. 2015, doi: 10.1080/01431161.2015.1054049.

[4] W. Wang *et al.*, "Pyramid Vision Transformer: A Versatile Backbone for Dense Prediction without Convolutions," *Proceedings of the IEEE International Conference on Computer Vision*, pp. 548–558, Feb. 2021, doi: 10.1109/ICCV48922.2021.00061.

[5] A. G. Howard *et al.*, "MobileNets: Efficient Convolutional Neural Networks for Mobile Vision Applications," Apr. 2017, Accessed: Oct. 03, 2024. [Online]. Available: http://arxiv.org/abs/1704.04861

[6] P. F. Rozario *et al.*, "Optimizing Mobile Vision Transformers for Land Cover Classification," *Applied Sciences 2024, Vol. 14, Page 5920*, vol. 14, no. 13, p. 5920, Jul. 2024, doi: 10.3390/APP14135920.

[7] X. X. Zhu *et al.*, "Deep Learning in Remote Sensing: A Comprehensive Review and List of Resources," *IEEE Geosci Remote Sens Mag*, vol. 5, no. 4, pp. 8–36, Dec. 2017, doi: 10.1109/MGRS.2017.2762307.

[8] W. Li, H. Fu, L. Yu, and A. Cracknell, "Deep Learning Based Oil Palm Tree Detection and Counting for High-Resolution Remote Sensing Images," *Remote Sensing 2017, Vol. 9, Page 22*, vol. 9, no. 1, p. 22, Dec. 2016, doi: 10.3390/RS9010022.

[9] F. Jiang *et al.*, "Artificial intelligence in healthcare: past, present and future," *Stroke Vasc Neurol*, vol. 2, no. 4, pp. 230–243, Dec. 2017, doi: 10.1136/SVN-2017-000101.

[10] A. Dosovitskiy *et al.*, "AN IMAGE IS WORTH 16X16 WORDS: TRANSFORMERS FOR IMAGE RECOGNITION AT SCALE," *ICLR 2021 - 9th International Conference on Learning Representations*, 2021.

[11] Y. Bazi, L. Bashmal, M. M. Al Rahhal, R. Al Dayil, and N. Al Ajlan, "Vision Transformers for Remote Sensing Image Classification," *Remote Sensing 2021, Vol. 13, Page 516*, vol. 13, no. 3, p. 516, Feb. 2021, doi: 10.3390/RS13030516.

[12] D. Hong, L. Gao, J. Yao, B. Zhang, A. Plaza, and J. Chanussot, "Graph Convolutional Networks for Hyperspectral Image Classification," *IEEE Transactions on Geoscience and Remote Sensing*, vol. 59, no. 7, pp. 5966–5978, Jul. 2021, doi: 10.1109/TGRS.2020.3015157.

[13] S. Wang *et al.*, "A deep learning framework for remote sensing image registration," *JPRS*, vol. 145, pp. 148–164, Nov. 2018, doi: 10.1016/J.ISPRSJPRS.2017.12.012.

[14] B. Gonzalez-Moodie, S. Daiek, J. Lorenzo-Trueba, J. and A.S. Varde, "Multispectral drone data analysis on coastal

dunes". IEEE International Conference on Big Data, 2021 (pp. 5903-5905). IEEE.

[15] A. Prasad, A. S. Varde, R. Gottimukkala, C. Alo, and P. Lal, "Analyzing Land Use Change and Climate Data to Forecast Energy Demand for a Smart Environment". IEEE International Renewable and Sustainable Energy Conference (IRSEC), 2021 (pp. 1-6).

[16] J. E. Ball, D. T. Anderson, and C. S. Chan, "Comprehensive survey of deep learning in remote sensing: theories, tools, and challenges for the community," *J Appl Remote Sens*, vol. 11, no. 04, p. 1, Sep. 2017, doi: 10.1117/1.JRS.11.042609.

[17] A. Pampoore-Thampi, A.S. Varde, and D. Yu, "Mining GIS data to predict urban sprawl". ACM KDD 2014 Conference, Bloomberg Track, 2021 arXiv preprint arXiv:2103.11338,

[18] P. Basavaraju, and A. S. Varde, "Supervised learning techniques in mobile device apps for Androids". ACM SIGKDD Explorations, 2016, 18(2), pp.18-29.

[19] Z. Sun, L. Di, and H. Fang, "Using long short-term memory recurrent neural network in land cover classification on Landsat and Cropland data layer time series," *Int J Remote Sens*, vol. 40, no. 2, pp. 593–614, Jan. 2019, doi: 10.1080/01431161.2018.1516313.

[20] F. M. Suchanek, A.S. Varde, P. Nayak, and P. Senellart, "The hidden Web, XML and the semantic Web: Scientific data management perspectives". In Proceedings of the 14th International Conference on Extending Database Technology, ACM EDBT, 2011 (pp. 534-537).

[21] A. Garg, N. Tandon, and A. S. Varde, "I am guessing you can't recognize this: Generating adversarial images for object detection using spatial commonsense". Proceedings of the AAAI Conference on Artificial Intelligence, 2020 (Vol. 34, No. 10, pp. 13789-13790).

[22] A. Garg, N. Tandon, and A. S. Varde, "CSK-SNIFFER: Commonsense Knowledge for Sniffing Object Detection Errors". ACM EDBT/ICDT Workshops, 2022.

[23] B. Youssef *et al.*, "Federated Learning Approach for Remote Sensing Scene Classification," *Remote Sensing 2024, Vol. 16, Page 2194*, vol. 16, no. 12, p. 2194, Jun. 2024, doi: 10.3390/RS16122194.

[24] G. Coppola, A.S. Varde, and J. Shang, "Enhancing Cloud Security Posture for Ubiquitous Data Access with a Cybersecurity Framework Based Management Tool". In IEEE 14th Annual Ubiquitous Computing, Electronics & Mobile Communication Conference (UEMCON), 2023, (pp. 0590-0594).

[25] A. Kaluarachchi, D. Roychoudhury, A. S. Varde, and G. Weikum, G., "SITAC: discovering semantically identical temporally altering concepts in text archives". Proceedings of the 14th International Conference on Extending Database Technology, ACM EDBT, 2011 (pp. 566-569).

[26] N. Tandon, A. S. Varde, and G. de Melo, "Commonsense knowledge in machine intelligence". ACM SIGMOD Record, 2017, 46(4), pp.49-52.

[27] S. Razniewski, N. Tandon, and A.S. Varde, "Information to wisdom: Commonsense knowledge extraction and compilation". Proceedings of the 14th ACM International Conference on Web Search and Data Mining, ACM WSDM, 2021 (pp. 1143-1146).

[28] C. J. Conti, A. S. Varde, and W. Wang, "Robot action planning by commonsense knowledge in human-robot collaborative tasks. IEEE International IOT, Electronics and Mechatronics Conference (IEMTRONICS), 2020 (pp. 1-7). IEEE.

[29] R. Hidalgo, J. Parron, A.S. Varde, and W. Wang, "Robo-csk-organizer: Commonsense knowledge to organize detected objects for multipurpose robots", Springer IEMTRONICS conference, 2024. arXiv preprint arXiv:2409.18385.

[30] P. Persaud, A. S. Varde, and S. Robila, "Enhancing autonomous vehicles with commonsense: Smart mobility in smart cities". IEEE 29th international conference on tools with artificial intelligence (ICTAI), 2017 (pp. 1008-1012).

[31] A.S. Varde, "Challenging research issues in data mining, databases and information retrieval". ACM SIGKDD Explorations, 2009, 11(1), pp.49-52.

[32] L. Wang, "Heterogeneous data and big data analytics. Automatic Control and Information Sciences", 2017, 3(1), pp.8-15.

[33] W. Liang, P.D. Meo, Y. Tang, and J. Zhu, "A survey of multi-modal knowledge graphs: Technologies and trends", 2024. ACM Computing Surveys, 56(11), pp.1-41.